\documentclass{edm_article}
\usepackage{graphicx}
\usepackage{booktabs}
\usepackage{url}
\usepackage{multirow}
\begin{document}

\title{Automated Visual Attention Detection using Mobile Eye Tracking in Behavioral Classroom Studies
}

\numberofauthors{4}
\author{
\alignauthor
Efe Bozkir\\
        \affaddr{Technical University of Munich}\\
        \affaddr{Arcisstr. 21}\\
        \affaddr{80333, Munich, Germany}\\
        \email{efe.bozkir@tum.de}
\alignauthor
Christian Kosel\\
        \affaddr{Technical University of Munich}\\
        \affaddr{Arcisstr. 21}\\
        \affaddr{80333, Munich, Germany}\\
        \email{christian.kosel@tum.de}
\alignauthor 
Tina Seidel\\
        \affaddr{Technical University of Munich}\\
        \affaddr{Arcisstr. 21}\\
        \affaddr{80333, Munich, Germany}\\
        \email{tina.seidel@tum.de}
\and
\alignauthor 
Enkelejda Kasneci\\
        \affaddr{Technical University of Munich}\\
        \affaddr{Arcisstr. 21}\\
        \affaddr{80333, Munich, Germany}\\
        \email{enkelejda.kasneci@tum.de}
}

\maketitle

\begin{abstract}
Teachers' visual attention and its distribution across the students in classrooms can constitute important implications for student engagement, achievement, and professional teacher training. Despite that, inferring the information about where and which student teachers focus on is not trivial. Mobile eye tracking can provide vital help to solve this issue; however, the use of mobile eye tracking alone requires a significant amount of manual annotations. To address this limitation, we present an automated processing pipeline concept that requires minimal manually annotated data to recognize which student the teachers focus on. To this end, we utilize state-of-the-art face detection models and face recognition feature embeddings to train face recognition models with transfer learning in the classroom context and combine these models with the teachers' gaze from mobile eye trackers. We evaluated our approach with data collected from four different classrooms, and our results show that while it is possible to estimate the visually focused students with reasonable performance in all of our classroom setups, U-shaped and small classrooms led to the best results with accuracies of approximately 0.7 and 0.9, respectively. While we did not evaluate our method for teacher-student interactions and focused on the validity of the technical approach, as our methodology does not require a vast amount of manually annotated data and offers a non-intrusive way of handling teachers' visual attention, it could help improve instructional strategies, enhance classroom management, and provide feedback for professional teacher development.
\end{abstract}

\keywords{educational technologies, eye tracking, computer vision, machine learning, deep learning}

\section{Introduction}
The integration of computer vision and machine learning techniques into the educational context has led to a new era of data analysis due to the likelihood of automating various time-consuming and non-trivial tasks that require manual human inputs and labeling~\cite{buehler_etal_2023,goldberg2021attentive}. Traditional educational research methods that focus on exploring teacher and student behavior in the classroom, relying heavily on human observation and self-reports such as surveys and interviews, have provided valuable insights into teacher and student behavior~\cite{Apter2010}. However, they face challenges in scalability, especially when analyzing behaviors across multiple classrooms in various settings and institutions. Moreover, even with trained human observers manually coding the data, it is possible to miss subtle behavioral nuances among diverse student groups. The advent of computer vision and machine learning techniques can address some of these challenges if such techniques work accurately. Some previous examples to this end include employing machine learning models to predict students' cognitive load~\cite{appel_etal_2023} from physiological and behavioral data in classrooms~\cite{yong2022recognizing} and analyzing students' learning-related behaviors~\cite{ABDALLAH2021951}. However, most approaches focus on one-sided attention, such as understanding students' behavioral patterns. These aspects can be combined and fused with teachers' attention and visual behavior simultaneously, providing the opportunity to have a fine-grained understanding of what happens in the classroom, and equipping teachers with eye trackers can enable this. 

Eye tracking has been utilized in different domains, such as education~\cite{Strohmaier2020,Busjahn_etal_2014}, medicine~\cite{HAREZLAK2018176,bradshaw_etal_2019}, marketing~\cite{wedel_and_pieters_2008,boerman_and_mueller_2022}, entertainment~\cite{ALMEIDA20161,moeller_etal_2021}, and driving~\cite{8797758,bozkir_etal_2019} and for different tasks including reading~\cite{suemer_etal_2021,bozkir_etal_2022}, problem-solving~\cite{TSAI2012375,LEE2019268}, and expertise assessment~\cite{castner_etal_2018,hosp_etal_2021}. Yet, many of these setups explicitly use remote and screen-based eye trackers, requiring the users to sit and remain stable during data collection, which might affect ecological validity negatively. When classroom context and especially teachers are considered, using mobile eye tracking (MET) might solve this issue to a certain extent while providing fine-grained visual attention data of teachers during the classroom discourse~\cite{Cortina_etal_2015}. 

In a broad sense, non-invasive eye tracking is widely recognized as one of the powerful methods in behavioral research for studying human cognition and perception~\cite{Eckstein2017}, while its popularity has grown in cognitive-oriented studies within educational psychology, including investigations into research on classroom observation and teacher expertise~\cite{jarodzka2021eye,Seidel2021}. As eye-tracking methods mostly capture gaze patterns and pupillometry, this provides insights into how teachers orient themselves in school classrooms that are characterized by a high level of social interaction and close proximity among students, resulting in a focused social environment that is highly relevant for study attention processes. To this end, previous research has focused on how teachers allocate their attention in the classroom, interact with students, and create effective learning environments~\cite{Grub}. By examining teachers' gaze patterns and visual attention, studies aimed to understand the dynamics of teacher-student interactions, student engagement, and classroom management~\cite{KOSEL2023104142,jarodzka2021eye,Sumer_2018_CVPR_Workshops}. 

Despite the aforementioned potentials in the classroom context, much of what is currently known about teachers' perception in the classroom is through lab-based and on-action eye-tracking studies in which participants observe classroom video sequences of others~\cite{Grub}. As a result, there is a need for systematic studies of teachers' perceptions in the classroom while teaching, not merely while observing~\cite{jarodzka2021eye}. MET is a powerful way of capturing teacher attention during in-action real classroom teaching without disrupting the instruction flow~\cite{Sumer_2018_CVPR_Workshops}. However, in many studies, MET data analysis is grounded in in-depth area-of-interest (AOI)-based analysis, either using time-consuming and error-prone frame-by-frame manual gaze mapping~\cite{Chukoskie,Vansteenkiste} or dynamic AOIs using either keyframes or different types of markers~\cite{Evans_Jacobs_Tarduno_Pelz_2012}. Both methods have drawbacks, especially when participants move and interact with objects and other people, which is often the case in the classroom context, making MET studies with several hours of recordings difficult to analyze quantitatively. As a result, researchers are reluctant to incorporate MET in their study designs~\cite{Vansteenkiste}. 

To overcome these methodological limitations and provide ecological validity, we developed a specialized processing pipeline to automatically map teacher attention on the students during classroom interactions using computer vision and machine learning techniques, to answer the research question (RQ) of ``How can novel face detection and recognition methods be integrated with mobile eye tracking to map teacher attention in conventional classroom settings?''. Answering this RQ, our work aims to empirically show that such a pipeline can accurately work and support further research on teacher-student interactions and behaviors, while supporting the research on teacher-student interactions. To this end, we first train face recognition models for the students in our classrooms using the very initial part of the scene videos of egocentric eye trackers by extending state-of-the-art face detection and recognition methods with transfer learning. Then, in the second part, using the estimated gaze point of the teacher, we identify the closest face patch to the teachers' gaze in the scene and feed this patch to our face recognition models to identify visually attended students for each frame in the egocentric video streams. As we manually annotated student face patches only from the first minute of the video data for our face recognition models, due to the effectiveness of our approach, in practice, practitioners would only need a minimal amount of student face data to automatically allocate the attention in video streams, without heavily relying on manual annotations. We show the effectiveness and accuracy of our technical solution in four different classroom situations, and to the best of our knowledge, our work is the first to conduct such analyses in authentic classrooms. Considering all, our major contributions are as follows: 

\begin{itemize}
  \item We propose a novel, automated computational processing pipeline to detect the student that the teacher visually focuses on in the conventional classroom context, achieving substantially better performance compared to chance by utilizing and further extending state-of-the-art face detection and recognition models (i.e., SCRFD-10GF~\cite{guo2022sample}, ResNet50 and ResNet100 models trained on WebFace600K~\cite{zhu2021webface260m} and Glint360K~\cite{an2021partialfctraining10} datasets) with conventional classifiers like random forest, support vector machine, and k-nearest neighbor by transfer learning. 
  \item We show that it is possible to accurately identify visually attended students in challenging setups, such as when eye-tracker scene cameras are used, where the teacher is non-stationary, and when the real-world environment includes partial occlusions and varying illumination conditions. 
  \item Our approach can provide further insights to facilitate research on the relationships between teachers' visual attention and different student characteristics, such as success, self-concept, and personality traits, while allowing more time for researchers to focus on the interpretation instead of dealing with manual annotations. 
\end{itemize}

\section{Related Work}
\label{sec_RW}
In this section, we discuss the related works in teacher attention studies using MET in classroom observation, and face processing and eye tracking in the wild in Sections~\ref{subsec_met_classroom} and~\ref{subsec_tech_rw_edu_tech}, respectively. 

\subsection{Eye-tracking Data in Classroom Studies}
\label{subsec_met_classroom}
Examining the quality of learning-relevant interactions between teachers and students in the classroom makes it possible to predict student achievement~\cite{inbook}. This relationship is supported by a large body of observational studies highlighting the critical role of pedagogical and emotional guidance in developing a student's cognitive, social, and emotional skills~\cite{POLING}. Previous works often employed various classroom observation methods (e.g., protocols, videotaped lessons, etc.) to explore the relationship between learning outcomes and classroom interactions. In this process, specially trained coders assess classroom dynamics, either live or through video recordings~\cite{9613750}. These coders assess various aspects, such as the teachers' responsiveness or the quality of feedback~\cite{Blomberg2013,Hofkens_etal2023}. In addition to traditional and high-inferential data obtained through manual coding of classroom interactions, researchers have begun collecting low-inferential data, such as eye tracking, to identify qualitative differences in teacher behaviors and cognition~\cite{Grub}. However, most studies in this area relied on on-action and lab-based eye-tracking data~\cite{Grub}. These studies often show participants video sequences from authentic teaching situations and analyze their attention regarding different aspects of classroom interactions, such as student engagement and disruptions~\cite{KOSEL2023104142}. 

In recent years, the availability of cost-effective glasses-based MET has enhanced the study of gaze behavior beyond the lab. MET glasses include two types of integrated cameras: an eye-tracking camera positioned under the eye and a scene camera typically mounted above the eyebrows on the frame~\cite{tobii_pro_glasses2_manual_2016}. Through precise calibration, the glasses combine the measured gaze point captured by the eye-tracking camera with the broader perspective provided by the scene camera. This integration allows for a comprehensive analysis of an individual's visual attention and facilitates a deeper understanding of their gaze behavior in the wild~\cite{7891939,Sumer_2018_CVPR_Workshops}. However, only few studies conducted in classrooms with teachers have used MET~\cite{Cortina2018,huang2021teachers,mcintyre2019capturing,goldberg2021attentive,pouta2021student}.

To this end, Goldberg et al.~\cite{goldberg2021attentive} analyzed which learners' behavior attracted pre-service teachers' attention during instruction using MET. The authors synchronized eye-tracking data with a continuous annotation of observable learning behavior and found that pre-service teachers' attention was mainly attracted by students' salient on-task learning-related behaviors, such as asking questions, raising their hands, or reflecting on something out loud. Pouta et al.~\cite{pouta2021student} went beyond merely recording frontal teaching, a common focus in previous studies~\cite{Cortina2018}. Instead, they honed in on specific episodes of student-teacher interactions. The authors analyzed the teachers' visual perception during these interactions, examining their gaze directions. This approach allowed them to distinguish between passive observation and active engagement.

Teachers' visual attention can rest on students without immediate engagement, serving as a background monitoring process~\cite{KOSEL2023104142}. Research showed that these situations, in which teachers monitor students, are used for collecting information about students' levels of interest and motivation and constitute a large part of their visual behavior in the classroom. Researchers are particularly interested in this monitoring behavior because it involves more top-down driven gaze processes, reflecting deliberate and goal-oriented attention allocation by teachers. At other times, teachers direct their gaze in response to specific student behaviors, such as a raised hand or a verbal question, engaging with students in meaningful ways~\cite{goldberg2021attentive}. Understanding how teachers navigate between these different attentional states underscores the complexity of their interactions with students and highlights the importance of context in interpreting gaze data. Previous research also recognizes this complexity for future studies~\cite{Grub}. In the context of automated attention understanding, a system should also work accurately in situations of passive observation, where the face of the student is not always perfectly visible from the front, as is more often the case in situations of direct interaction or hand-raising. When such a system can accurately detect and interpret visual attention in various classroom scenarios, it will likely provide a more comprehensive understanding of teacher-student interactions. It is then also possible to further extend such analyses towards a multi-modal fashion in more meaningful ways, as previous works partly addressed by extracting teaching activities from the sensor data~\cite{Prieto_etal_2018}, analyzing expertise for teaching situations~\cite{gao_etal_ismar_23}, and capturing the processes of social interaction~\cite{hannula_etal_2022}. 

While there has been some progress in understanding teachers' attention processes in the wild, our knowledge remains limited~\cite{jarodzka2021eye}. The studies in this context suffer from significant constraints that hinder a comprehensive understanding, as many of them are characterized by a small sample size, which can reduce the generalizability of the findings~\cite{goldberg2021attentive}. Additionally, the limited recording time in these investigations may not capture the full spectrum of attentional behaviors~\cite{KOSEL2023104142}. These limitations highlight the need for more comprehensive and rigorous research to delve deeper into teachers' attentional processes~\cite{jarodzka2021eye}. There is a growing consensus, as highlighted by previous studies~\cite{goldberg2021attentive,suemer_2020_automated,buehler_etal_2023}, on the increasing need for automated procedures, especially in data handling and analysis. Combining computer vision and machine learning techniques in the context of face detection, face recognition, and eye tracking could offer valuable solutions for this purpose. 

\subsection{Face Processing and Eye Tracking In the Wild}
\label{subsec_tech_rw_edu_tech}
Face detection and recognition tasks, particularly in the wild, are often considered challenging due to varying environmental factors such as illumination conditions, partial occlusions, and different camera viewing angles~\cite{ZAFEIRIOU20151,8600370}. Despite these, deep learning-based solutions provide decent performance even in such challenging conditions, providing the possibility of carrying out further analyses taking face processing results as bases~\cite{Ghazi_2016_CVPR_Workshops,WANG2021215}. The importance of face detection and processing in the wild was constantly emphasized~\cite{Masi_2016_CVPR,8600370,10472479,Zhao_2018_CVPR,jurevivcius2019method}, which is also essential for accurate educational data mining in the classroom context.

To this end, different approaches exist in the literature. Zhang et al.~\cite{faceboxes_ijcb_2017} proposed a face detector consisting of rapidly digested and multiple-scale convolutional layers that can run on CPU in real time, whose speed is invariant of detected faces. In another work, Li et al.~\cite{dsfd_cvpr_2019} introduced a novel face detection network with three key contributions, including a feature enhancer, adoption of a progressive anchor loss, and integration of an anchor assign strategy into data augmentation, achieving superior performance in several face detection benchmarks. Later, Deng et al.~\cite{retinaface_cvpr_2020} proposed RetinaFace, a single-shot and multi-level face localization method that unifies face box prediction, 2D facial landmark detection, and 3D vertices under the target of point regression on the image plane, achieving a stable face detection performance while keeping the efficiency through single-shot inference. Liu et al.~\cite{hambox_cvpr_2020} studied anchors in the context of face detection and observed that some unmatched anchors have a significant regression ability and further proposed an anchor mining strategy to sample high-quality anchors for the training phase. Such a mining strategy outperformed state-of-the-art methods in multiple datasets. 

Recently, Zhu et al.~\cite{zhu2021tinaface} treated the face detection problem as a generic object detection problem by using a ResNet50 backbone and showed in a single-model and single-scale setting that their method, TinaFace, outperforms the state-of-the-art, especially with test-time data augmentation. Guo et al.~\cite{guo2022sample} identified efficient face detection with low computation cost and high precision as an unsolved problem and focused on training data sampling and computation distribution strategies by introducing two methods, including a data augmentation technique for training samples in the most needed stages and a computation redistribution strategy by reallocating the computation between the backbone and head of the model. The proposed method, SCRFD, outperformed TinaFace~\cite{zhu2021tinaface} while being three times faster. 

Another essential step, especially for studies in the wild in the classroom context, is facial recognition in challenging conditions and with noisy data. To this end, Deng et al.~\cite{Deng_2019_CVPR} introduced an additive margin loss (i.e., ArcFace) and further extended ArcFace by sub-centering to deal with faces under real-world noise. The authors showed that without training any additional generator or discriminator network, the ArcFace can provide discriminative feature embeddings. While ArcFace provides effective feature embeddings, it assumes clean training data, and to address this issue and enable robustness to noise, Deng et al.~\cite{deng2020subcenter} further iterated ArcFace by introducing sub-classes for each identity and showed with large-scale raw web faces that their method achieves state-of-the-art face recognition performance in several benchmarks. In another work, Deng et al.~\cite{VPL_cvpr2021} proposed representing every class as a distribution instead of points in the latent space for face recognition, and the authors showed that their approach could simulate sample-to-sample comparisons for classification while being computationally efficient and memory-saving. Other work~\cite{zhu2021webface260m,an2022pfc} also tried to unveil the power of large-scale noisy datasets to improve the performance of face recognition models, yet acknowledged the rise in computing costs when training sets include a significant number of identities. 

While face detection and recognition tasks can be combined semantically to understand what is going on in the real-world environment, to understand the activities of users (i.e., teachers or students in the educational context) and how they react to a particular situation or stimulus, eye-tracking data can be helpful~\cite{jarodzka2021eye}. Fine-grained eye-tracking data requires dedicated hardware, such as MET devices, to estimate participants' gaze. The MET algorithms are different from the remote eye tracking setups, as in MET, eye cameras are in close proximity to the eyes. To tackle this challenge, Feng et al.~\cite{9756796} focused on a real-time gaze tracking algorithm that can operate on a mobile processor by continuously utilizing predicted regions of interests of near-eye images, and the authors achieved accuracies lower than one degree. Angelopoulos et al.~\cite{9389490} also focused on high-speed gaze tracking utilizing an event-based method offering eye-tracking frequencies beyond 10KHz by using an online 2D pupil fitting method. In another work, Lu et al.~\cite{9995692} proposed a pupil localization method with a novel deep learning-based corneal refraction correction and argued that pupil localization in 3D space is an essential intermediate algorithmic step for accurate gaze tracking and showed that their method helps significantly decrease gaze estimation error. 

Ranjan et al.~\cite{Ranjan_2018_CVPR_Workshops} focused on an appearance-based and head-pose invariant gaze estimation using convolutional neural networks and showed that their approach is ten times faster than the state-of-the-art approaches. Similarly, Palmero et al.~\cite{palmero_etal_2020} and Stojanov et al.~\cite{Stojanov_etal_2022} focused on the benefits of temporal and depth information on the gaze estimation tasks and found that incorporating both temporal and depth information helps to improve the performance of gaze estimation tasks. Other work focused on unsupervised representation learning to learn low-dimensional gaze representations without gaze annotations~\cite{Yu_2020_CVPR} and the notion of two-eye asymmetry to utilize an asymmetric regression network to predict 3D gaze directions~\cite{Cheng_2018_ECCV}. While various works focused on different aspects of gaze estimation, face recognition, and face detection, they constantly emphasize the importance of real-time working capability, robustness to noise, and the efficacy of the methods for the real-world use of algorithms. We took these recommendations into account while selecting the methods we use for the education context in the classrooms and utilized SCRFD-10GF~\cite{guo2022sample} for the face detection and the ResNet50 architecture that was trained on the WebFace600K dataset (ResNet50@WebFace600K)~\cite{zhu2021webface260m} for face recognition. We benchmarked facial embeddings of ResNet50@WebFace600K against facial embeddings that come from a ResNet100 trained on the Glint360K dataset (ResNet100@Glint360K)~\cite{an2021partialfctraining10}, which is considered one of the largest and cleanest face recognition datasets in the literature. Upon obtaining optimal facial embeddings, we utilized transfer learning with several classifiers to identify visually attended students by leveraging facial embeddings and teachers' gaze in the classroom context.

\section{Methodology}\label{sec_Methods}
Our methodology has two main pillars, including the MET data collection and automated teacher attention mapping. We discuss these separately in Sections~\ref{subsec_experimental_design} and~\ref{subsec_auto_attention_mapping}, respectively. 

\subsection{Experimental Design}
\label{subsec_experimental_design}
The following subsections introduce our experimental design, particularly our participants, data types, and data collection procedure. The Bavarian Ministry of Education and Cultural Affairs reviewed and approved our data protection measures, the design and necessity of the survey, and ethical points. 

\subsubsection{Participants and Data}
Four in-service mathematics teachers (two women and two men) participated in the data collection. Teachers were between 27 and 62 years old ($M = 37.25, SD = 16.64$). The class sizes ranged from 14 to 25 students (69 students total). The collected data included video streams from the classrooms and eye-tracking data (i.e., gaze data) recorded from the mobile eye trackers. The resolution of the videos recorded from the eye trackers' scene camera is $1920 \times 1080$. 

\subsubsection{Data Collection Procedure}
The mobile eye-tracking recordings took place during a regular class period, chosen to interfere as little as possible with the regular lesson plan. We used a Tobii Pro Glasses 2~\cite{tobii_pro_glasses2_manual} with an eye-tracking sampling frequency of 50 Hz to collect eye movement data. Recordings started after a successful one-point eye-tracking calibration. If an interruption due to a technical error happened, we recalibrated the eye-tracking system in the same way. During the data collection process, we connected the eye-tracking system to a computer via a wireless connection, where we controlled the recordings with the Tobii Pro Glasses Controller ($\times 64$)~\cite{tobii_pro_glasses2_manual_2016}. 

Before data collection began, all participants, including students, teachers, and the students' parents (or legal guardians), provided written informed consent. The consent forms explicitly outlined the scope and purpose of the study, ensuring that all parties were fully aware that the sessions would involve video recordings and eye-tracking technology. Participants also agreed to appear in academic publications or presentations in disguised forms, such as blurred or altered screenshots. 

\begin{figure*}[ht]
    \centering
    \begin{minipage}{0.46\textwidth}
        \centering
        \includegraphics[width=0.95\linewidth]{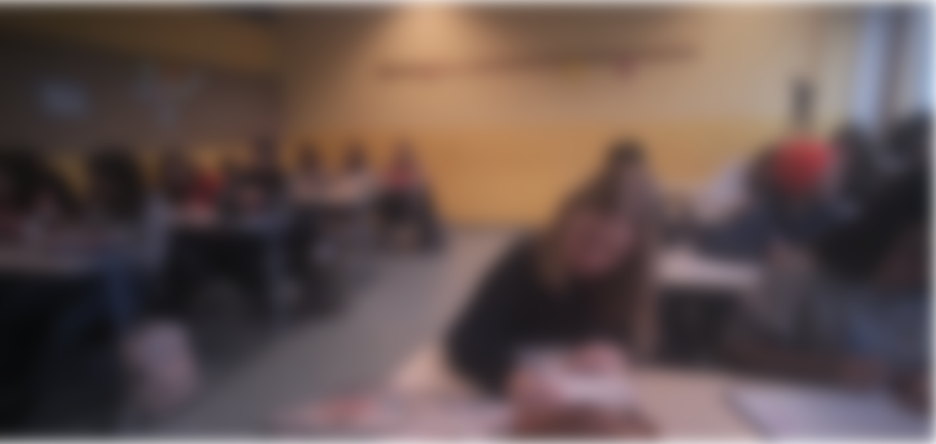}
        \textbf{(a) A view from the first classroom.}
    \end{minipage}
    \begin{minipage}{0.46\textwidth}
        \centering
        \includegraphics[width=0.95\linewidth]{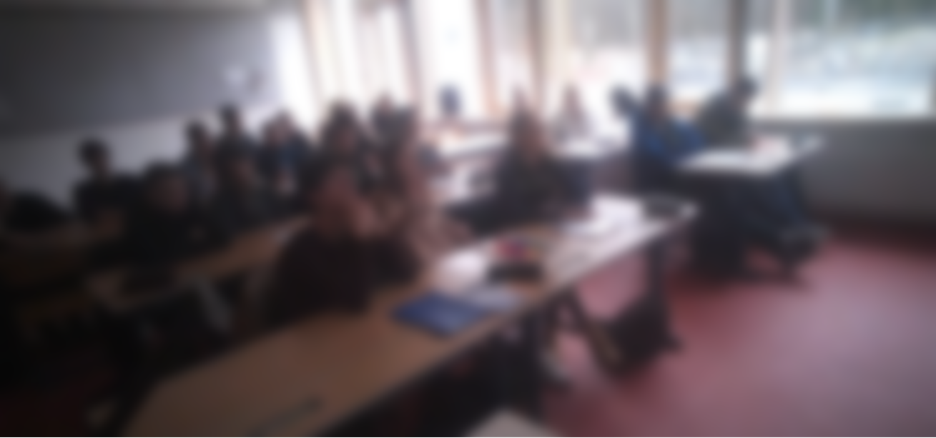}
        \textbf{(b) A view from the second classroom.}
    \end{minipage}
    \begin{minipage}{0.46\textwidth}
        \centering
        \includegraphics[width=0.95\linewidth]{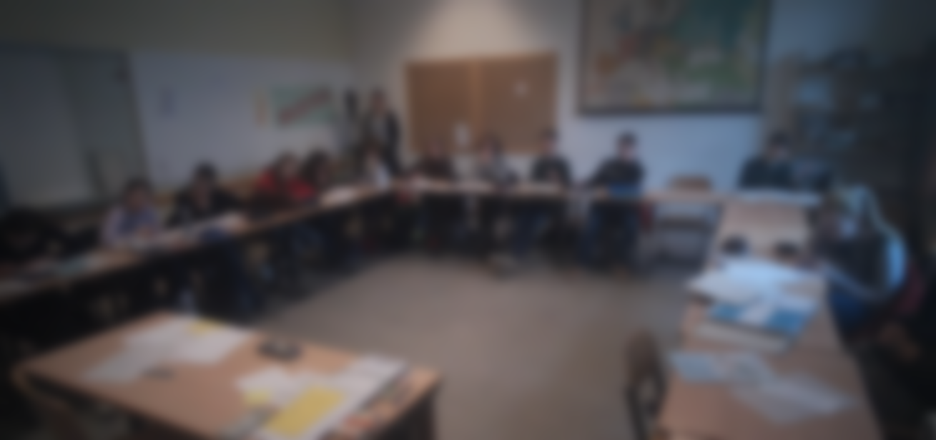}
        \textbf{(c) A view from the third classroom.}
    \end{minipage}
    \begin{minipage}{0.46\textwidth}
        \centering
        \includegraphics[width=0.95\linewidth]{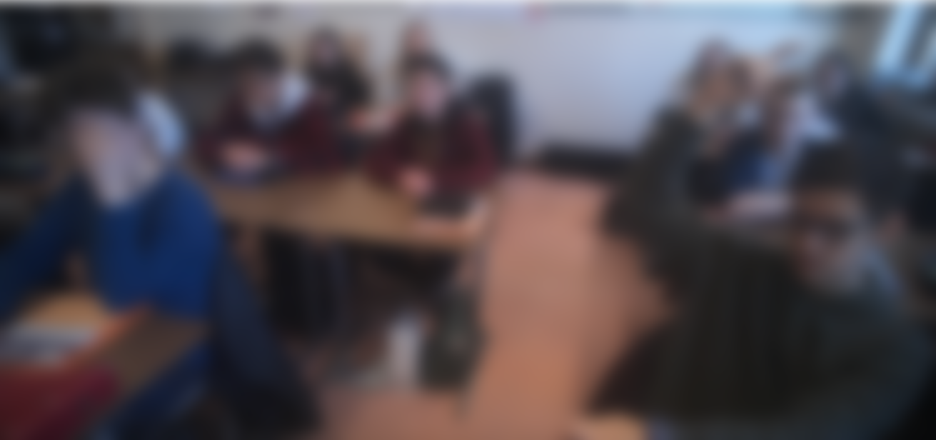}
        \textbf{(d) A view from the fourth classroom.}
    \end{minipage}
    \caption{Views from the video recordings from eye trackers that teachers wear in each of our four classrooms. We provide blurred versions of the images to protect students' privacy.}
    \label{fig_classroom_screenshots}
\end{figure*}

After the consent phase, each teacher gave a lesson ranging between 60 and 70 minutes in four different higher secondary schools (ninth grade) in Germany. All participating teachers taught similar content (i.e., matrix calculus) during the data collection. In addition, the sampled lesson was minimally pre-determined to allow for consistency across teachers and their lessons. The teachers had five minutes to recap the topic and tasks of the last lesson and the remaining time to introduce new content. We advised the participating teachers to keep their eye-tracking glasses stable throughout the lesson. At the end of the data collection procedure, we thanked teachers and students for participating in our study. Figure~\ref{fig_classroom_screenshots} depicts different classrooms from which we collected the data. 

\subsection{Automated Attention Mapping of Teacher}
\label{subsec_auto_attention_mapping}
We first introduce our general methodology on automated attention mapping of teachers in Section~\ref{subsec_method}. We then provide implementation and evaluation details in Section~\ref{subsec_implementation}. 

\subsubsection{Method}
\label{subsec_method}
\begin{figure*}[ht]
    \centering
        \centering
        \includegraphics[width=0.97\textwidth]{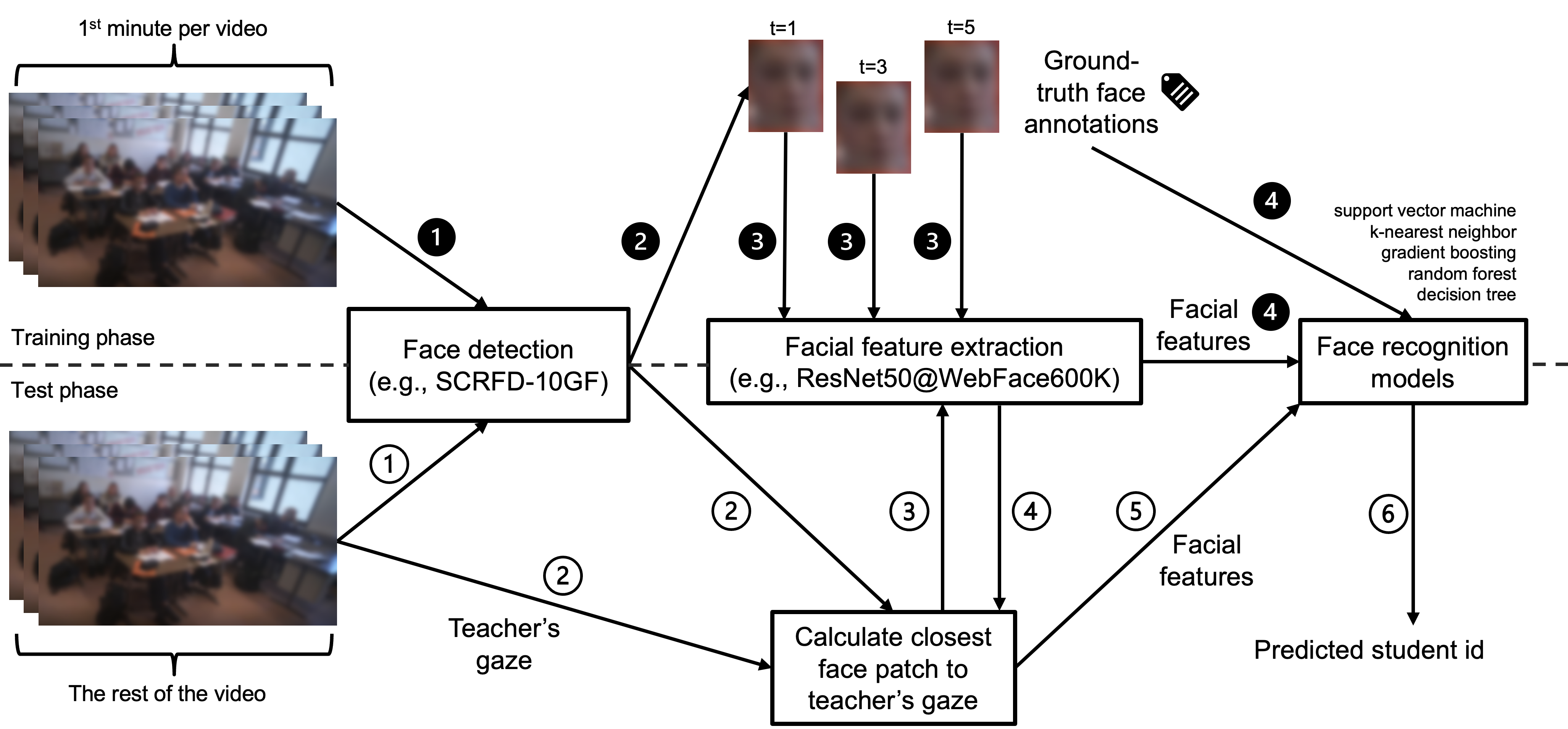}
    \caption{Overview of our automated teacher attention mapping pipeline. We provide blurred versions of the images to protect student privacy. Dashed lines separate the training and testing processes, whereas numbers represent the operations' order. We use Euclidean distance when calculating the closest patch to the teacher's gaze.}
    \label{fig_processing_pipeline}
\end{figure*}

The primary purpose of the proposed automated attention mapping pipeline is to automatically identify which student teachers visually focus on at each time point, and this processing pipeline includes two main components. We explain those in the following and summarize the overall process in Figure~\ref{fig_processing_pipeline}. The results of this attention mapping pipeline can be used as a baseline for further student-teacher interaction analysis or educational data mining in the classroom context.

\textbf{\textit{Face detection and recognition.}} To identify which student is visually focused on by the teacher, as a first step \textbf{(i)}, the use of face detection and recognition models is necessary. For our pipeline, we selected the SCRFD-10GF~\cite{guo2022sample} and the ResNet50@WebFace600K models~\cite{zhu2021webface260m} for face detection and recognition, respectively. SCRFD-10GF includes a substantially smaller number of model parameters and comes with a shorter inference time compared to various other methods for face detection~\cite{zhu2021tinaface,retinaface_cvpr_2020}. Considering that a smaller number of parameters and shorter inference times for face detection will also help with real-time face detection through the scene cameras of the eye tracker if deployed, the use of SCRFD-10GF is reasonable, as it also achieves good performance in the face detection tasks, even in challenging setups. As for the face recognition part, to obtain facial embeddings, we use the ResNet50 architecture that was trained on the WebFace600K dataset, which was cleaned in an automated fashion and could include noisy samples, similar to our face instances from classroom environments. We also apply face alignment with five facial key points (i.e., eye centers, nose, and mouth corners) between our face detection and facial embedding components.

While the aforementioned publicly available machine learning models were trained and evaluated with many users, they do not include students from our classroom setups. This is not an issue for face detection tasks, but for face recognition, we need to incorporate our individuals and their data from the classrooms into our face recognition pipeline. To this end, instead of carrying out the face recognition task with the ResNet50@WebFace600K model, we use its facial feature embeddings of size $512$ for transfer learning and further train five conventional classifiers, including the random forest, k-nearest neighbor, support vector machine, gradient boosting, and decision tree classifiers. These models serve as baselines for the teachers' attention mapping. 

\textbf{\textit{Teacher gaze mapping.}} As the second step in our pipeline \textbf{(ii)}, mapping teachers' gaze to particular students is necessary to complete the attention mapping pipeline. In this step, we rely on the estimated gaze from the eye tracker on video recordings. In parallel, we apply face detection via SCRFD-10GF on the frames coming from the eye tracker's scene camera. For each frame, once our processing pipeline provides detected faces and 2D coordinates of the gaze point of the teacher, we identify the closest detected face patch to the teacher's gaze. To identify the closest face patch, we use the Euclidean distance between the teacher's gaze and the center point of each face patch on the image frames. We do not consider teachers' eye movement type (e.g., fixation, saccade) and keep this aspect exploratory. Following, we send the feature embedding of the closest face patch, which is extracted from the pre-trained ResNet50@WebFace600K, to our face recognition model (i.e., random forest, support vector machine, etc.) to identify which student the teacher visually focuses on. We carry out this process continuously during the lesson while the teacher wears MET. 

\subsubsection{Implementation and Evaluation}
\label{subsec_implementation}
We implemented our attention mapping processing pipeline using InsightFace~\cite{insightface_lib} and scikit-learn~\cite{scikit_learn_lib} libraries in Python and evaluated it for each classroom video separately, in an offline fashion. To train the face recognition models, we first applied our face detection pipeline frame by frame to each video. We randomly annotated 30 detected face instances per student from the first minute of the videos on the detected faces, and we decided on this number empirically from our data. We used the facial feature embeddings extracted from those 30 face instances from the ResNet50@WebFace600K model to train the random forest, k-nearest neighbor, support vector machine, gradient boosting, and decision tree classifiers to facilitate face recognition in the classroom context. For each of these classifiers, we applied 5-fold cross-validation to tune the hyperparameters of the classifiers by using a grid search approach. During the hyperparameter tuning phase, we preserved the percentages of samples for each class when splitting the data. We provide the complete set of hyperparameters in Table~\ref{tab_hyper_grid}. We used the tuned hyperparameters to train the final models for each classroom video for the evaluations. For feature normalization, we standardized features by removing the mean and scaling them to unit variance using the training data statistics to avoid data leakage in the validation and test phases. 

\begin{table}[ht!]
    \caption{Hyperparameter search grids for machine learning models. RF, SVM, k-NN, GB, and DT indicate random forest, support vector machine, k-nearest neighbor, gradient boosting, and decision tree classifiers, respectively. If we found the same best working hyperparameter for all the classrooms for each facial feature embedding type, we report them with bold values for ResNet50@WebFace600K, whereas with underlines for ResNet100@Glint360K.}
    \label{tab_hyper_grid}
    \begin{tabular}[t]{c|l}
        \toprule
        Model & Hyperparameter grid\\
        \midrule
        \multirow{6}{*}{RF} & Maximum depth: [10, 20, 40, 80, 160]\\
            & Maximum features: [sqrt, log2]\\
            & Minimum samples leaf: [\underline{1}, 2, 3, 4, 5]\\
            & Minimum samples split: [2, 4, 8, 16]\\
            & Number of estimators: [25, 50, 100, 200, 400, \\ 
            & 600, 800, 1000, 1200, 1400, 1600]\\
        \midrule
        \multirow{3}{*}{SVM} & C: [0.001, 0.005, 0.01, 0.05, 0.1, 0.5, 1]\\    
            & Gamma: [\textbf{\underline{auto}}, scale] \\
            & Kernel: [rbf, \underline{linear}, poly, sigmoid] \\
        \midrule
        \multirow{1}{*}{k-NN} & k: [\textbf{5}, 7, 9, 11, 13, 15, 17, 19, 21]\\
        \midrule
        \multirow{3}{*}{GB} & Number of estimators: [100, \underline{200}, 400, 600]\\
           & Minimum samples leaf: [1, \textbf{2}]\\    
           & Minimum samples split: [2, 4, \textbf{8}]\\
        \midrule
        \multirow{5}{*}{DT} & Maximum depth: [10, 20, 40, 80, 160]\\
           & Maximum features: [\textbf{sqrt}, log2]\\
           & Minimum samples leaf: [1, 2, 3, 4, 5]\\
           & Minimum samples split: [2, 4, 8, 16]\\
           & Criterion: [gini, entropy]\\
        \bottomrule
    \end{tabular}
\end{table}

We used the first minute of the videos to annotate students' faces because, at the beginning of the lessons, teachers' visual focus covers a greater field of view, making it possible to identify many student faces. In addition, while the activities that would happen in the later phases of the lessons can differ based on the classroom setup and lesson type, the first several minutes of the lessons are often generic in terms of activities that take place, which helps systematically apply and evaluate the designed processing pipeline. However, any ground truth annotation for the students' faces should help generate a similar processing pipeline. 

To implement gaze mapping and identify the visually attended student, while we relied on the gaze point Tobii eye trackers provided, we carried out some approximations in our automated pipeline to evaluate whether our pipeline works accurately. Firstly, for the gaze signal, if signals from both eyes were valid (i.e., gaze direction is detected accurately, no blinks occurred, etc.), we averaged the gaze locations of both eyes to have the final single point. If only one eye's signal was valid, we assumed that such information would also provide valuable insights into the teacher's attention and took the final gaze point on the classroom video from the single eye. Secondly, we calculated the center points of each face patch coming from the SCRFD-10GF face detection model and measured the Euclidean distances between these center points and the final gaze location on the same frame. If multiple faces were detected, we identified the focused student's face patch based on the shortest distance. We then fed the facial feature embeddings of the identified face patch based on the shortest distance to our classification models (e.g., random forest) and obtained the final result on the student ID. If no gaze signal was available due to both eyes giving invalid signals, we did not evaluate these cases, as a missing gaze signal would directly lead to a wrong result. 

To evaluate the gaze mapping part of our pipeline, we needed to have annotated ground truth data about where the teachers focus in the scene, considering the detected faces. Similar to face annotations from the first minute, we annotated approximately the second minute of each video with the information on which student teachers focused and conducted our evaluations according to this manually annotated ground-truth data. Our annotation process involved two trained annotators, and to ensure the reliability of the annotations, we established inter-rater reliability by calculating the Cohen’s kappa coefficient, which resulted in a high level of agreement ($\kappa= .94$), indicating consistency among annotators. We performed double coding for 25\% of the data. 

In evaluations, for facial feature embeddings, apart from the ResNet50@WebFace600K, we took ResNet100@Glint360K~\cite{an2021partialfctraining10} as a baseline since the Glint360K dataset includes cleaner face instances than our dataset, as well as the WebFace600K. The baseline evaluations with ResNet100@Glint360K only replace the ResNet50@WebFace600K component and keep the rest of the pipeline identical. Considering all, we evaluated our automated processing pipeline using data from four authentic classrooms, and in Table~\protect\ref{tbl_data_stats}, we report the number of training and test instances. 

\begin{table}[ht!]
\centering
\caption{Number of training and test samples for each classroom.}
\begin{tabular}{c|c|c}
  \toprule
  Classroom ID & \# Training Samples & \# Test Samples \\
  \midrule
  1 & 720 & 459 \\
  \hline
  2 & 510 & 502 \\
  \hline
  3 & 420 & 190 \\
  \hline
  4 & 390 & 935 \\
  \bottomrule
\end{tabular}
\label{tbl_data_stats}
\end{table}

\begin{table*}[ht!]
\centering
\caption{Results for automated teacher attention mapping. We indicate the best results and settings in bold for each classroom.}
\begin{tabular}{c|c|c|c|c|c|c}
\toprule
Classroom ID & Classifier & Facial Feature Embeddings & Accuracy & Precision & Recall & $\text{F}_1$ score \\
\midrule
\multirow{10}{*}{1} & \multirow{2}{*}{\textbf{Random Forest}} & \textbf{ResNet50@WebFace600K} & \textbf{0.50} & 0.68 & \textbf{0.50} & \textbf{0.56} \\
& & ResNet100@Glint360K & 0.46 & 0.64 & 0.46 & 0.52 \\
\cmidrule{2-7}
& \multirow{2}{*}{Support Vector Machine} & ResNet50@WebFace600K & 0.44 & 0.65 & 0.44 & 0.51 \\
& & ResNet100@Glint360K & 0.43 & 0.65 & 0.43 & 0.50 \\
\cmidrule{2-7}
& \multirow{2}{*}{k-Nearest Neighbor} & \textbf{ResNet50@WebFace600K} & 0.40 & \textbf{0.69} & 0.40 & 0.45 \\
& & ResNet100@Glint360K & 0.32 & 0.61 & 0.32 & 0.40 \\
\cmidrule{2-7}
& \multirow{2}{*}{Gradient Boosting} & ResNet50@WebFace600K & 0.24 & 0.42 & 0.24 & 0.28 \\
& & ResNet100@Glint360K & 0.25 & 0.46 & 0.25 & 0.29 \\
\cmidrule{2-7}
& \multirow{2}{*}{Decision Tree} & ResNet50@WebFace600K & 0.11 & 0.25 & 0.11 & 0.14 \\
& & ResNet100@Glint360K & 0.19 & 0.33 & 0.19 & 0.23 \\
\midrule
\multirow{10}{*}{2} & \multirow{2}{*}{\textbf{Random Forest}} & \textbf{ResNet50@WebFace600K} & \textbf{0.58} & 0.71 & \textbf{0.58} & \textbf{0.61} \\
& & ResNet100@Glint360K & 0.42 & 0.67 & 0.42 & 0.45 \\
\cmidrule{2-7}
& \multirow{2}{*}{Support Vector Machine} & ResNet50@WebFace600K & 0.49 & 0.72 & 0.49 & 0.54 \\
& & ResNet100@Glint360K & 0.40 & 0.73 & 0.40 & 0.47 \\
\cmidrule{2-7}
& \multirow{2}{*}{k-Nearest Neighbor} & \textbf{ResNet50@WebFace600K} & 0.44 & \textbf{0.73} & 0.44 & 0.48 \\
& & ResNet100@Glint360K & 0.18 & 0.59 & 0.19 & 0.23 \\
\cmidrule{2-7}
& \multirow{2}{*}{Gradient Boosting} & ResNet50@WebFace600K & 0.18 & 0.53 & 0.18 & 0.21 \\
& & ResNet100@Glint360K & 0.27 & 0.47 & 0.27 & 0.28 \\
\cmidrule{2-7}
& \multirow{2}{*}{Decision Tree} & ResNet50@WebFace600K & 0.17 & 0.42 & 0.17 & 0.23 \\
& & ResNet100@Glint360K & 0.12 & 0.38 & 0.12 & 0.15 \\
\midrule
\multirow{10}{*}{3} & \multirow{2}{*}{\textbf{Random Forest}} & \textbf{ResNet50@WebFace600K} & \textbf{0.68} & \textbf{0.74} & \textbf{0.68} & \textbf{0.69} \\
& & ResNet100@Glint360K & 0.60 & 0.62 & 0.60 & 0.59 \\
\cmidrule{2-7}
& \multirow{2}{*}{Support Vector Machine} & ResNet50@WebFace600K & 0.63 & 0.72 & 0.63 & 0.65 \\
& & ResNet100@Glint360K & 0.58 & 0.63 & 0.58 & 0.58 \\
\cmidrule{2-7}
& \multirow{2}{*}{k-Nearest Neighbor} & ResNet50@WebFace600K & 0.63 & 0.70 & 0.63 & 0.64 \\
& & ResNet100@Glint360K & 0.57 & 0.60 & 0.57 & 0.57 \\
\cmidrule{2-7}
& \multirow{2}{*}{Gradient Boosting} & ResNet50@WebFace600K & 0.32 & 0.63 & 0.32 & 0.36 \\
& & ResNet100@Glint360K & 0.38 & 0.55 & 0.38 & 0.41 \\
\cmidrule{2-7}
& \multirow{2}{*}{Decision Tree} & ResNet50@WebFace600K & 0.33 & 0.46 & 0.33 & 0.35 \\
& & ResNet100@Glint360K & 0.16 & 0.19 & 0.16 & 0.16 \\
\midrule
\multirow{10}{*}{4} & \multirow{2}{*}{Random Forest} & ResNet50@WebFace600K & 0.91 & 0.93 & 0.92 & 0.92 \\
& & ResNet100@Glint360K & 0.92 & 0.93 & 0.92 & 0.92 \\
\cmidrule{2-7}
& \multirow{2}{*}{Support Vector Machine} & \textbf{ResNet50@WebFace600K} & 0.94 & \textbf{0.95} & 0.94 & 0.94 \\
& & ResNet100@Glint360K & 0.93 & 0.94 & 0.93 & 0.93 \\
\cmidrule{2-7}
& \multirow{2}{*}{\textbf{k-Nearest Neighbor}} & \textbf{ResNet50@WebFace600K} & \textbf{0.95} & \textbf{0.95} & \textbf{0.95} & \textbf{0.95} \\
& & ResNet100@Glint360K & 0.89 & 0.92 & 0.89 & 0.90 \\
\cmidrule{2-7}
& \multirow{2}{*}{Gradient Boosting} & ResNet50@WebFace600K & 0.61 & 0.79 & 0.61 & 0.67 \\
& & ResNet100@Glint360K & 0.72 & 0.83 & 0.72 & 0.76 \\
\cmidrule{2-7}
& \multirow{2}{*}{Decision Tree} & ResNet50@WebFace600K & 0.51 & 0.68 & 0.51 & 0.56 \\
& & ResNet100@Glint360K & 0.31 & 0.64 & 0.31 & 0.39 \\
\bottomrule
\end{tabular}
\label{table_stats_results}
\end{table*}

\section{Results}\label{sec_Results}
\begin{figure*}[ht]
    \centering
    \begin{minipage}{0.49\textwidth}
        \centering
        \includegraphics[width=1\linewidth]{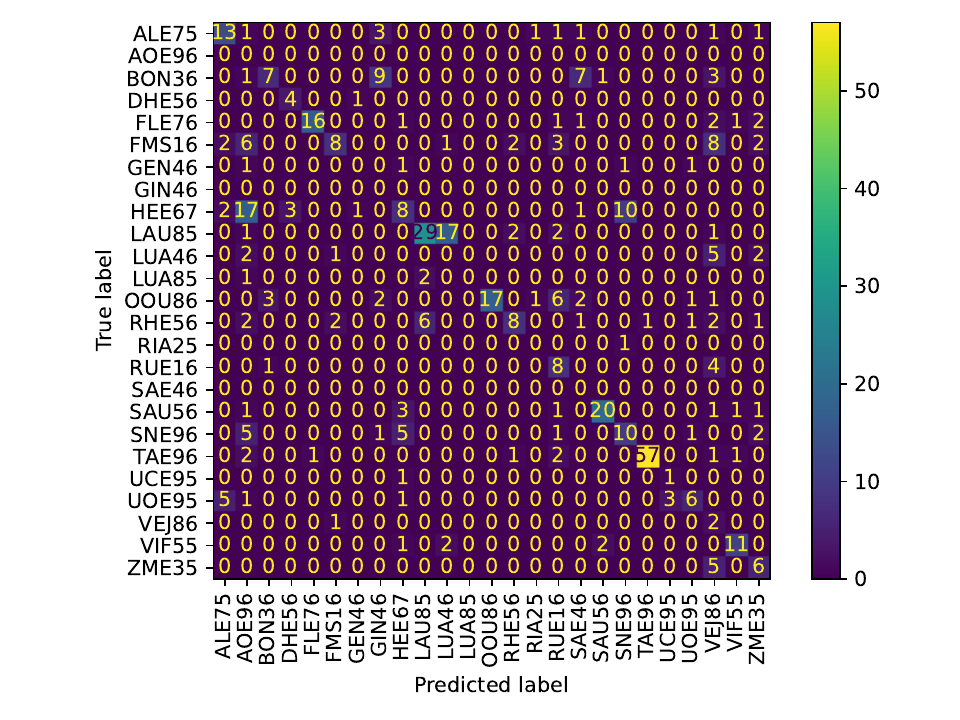}
        \textbf{(a) Confusion matrix of the random forest classifier for the first classroom.}
    \end{minipage}
    \begin{minipage}{0.49\textwidth}
        \centering
        \includegraphics[width=1\linewidth]{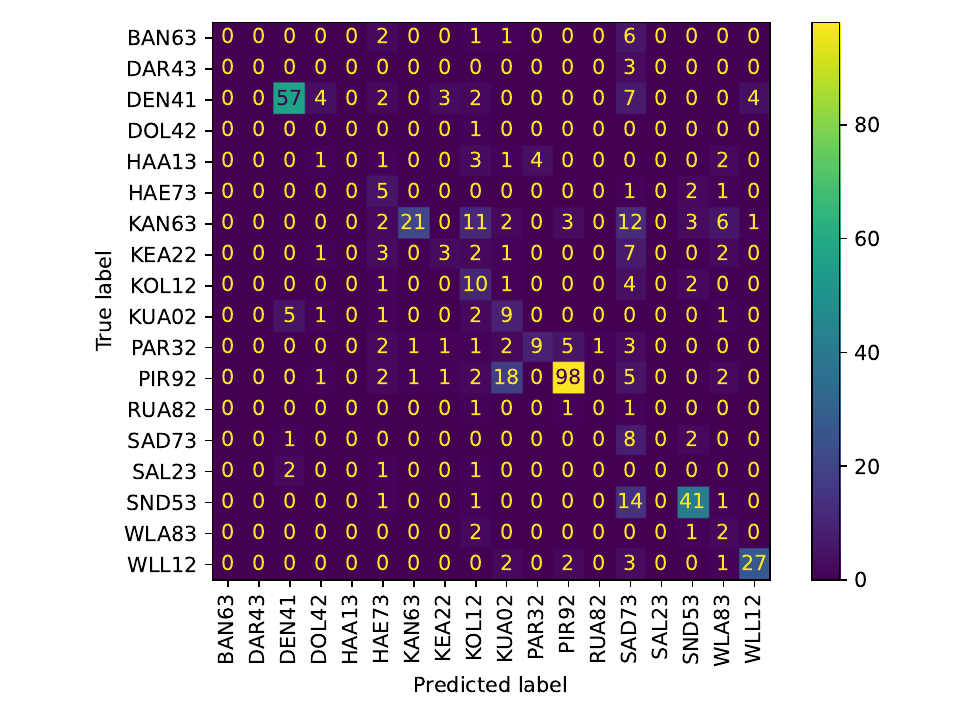}
        \textbf{(b) Confusion matrix of the random forest classifier for the second classroom.}
    \end{minipage}
    \begin{minipage}{0.49\textwidth}
        \centering
        \includegraphics[width=1\linewidth]{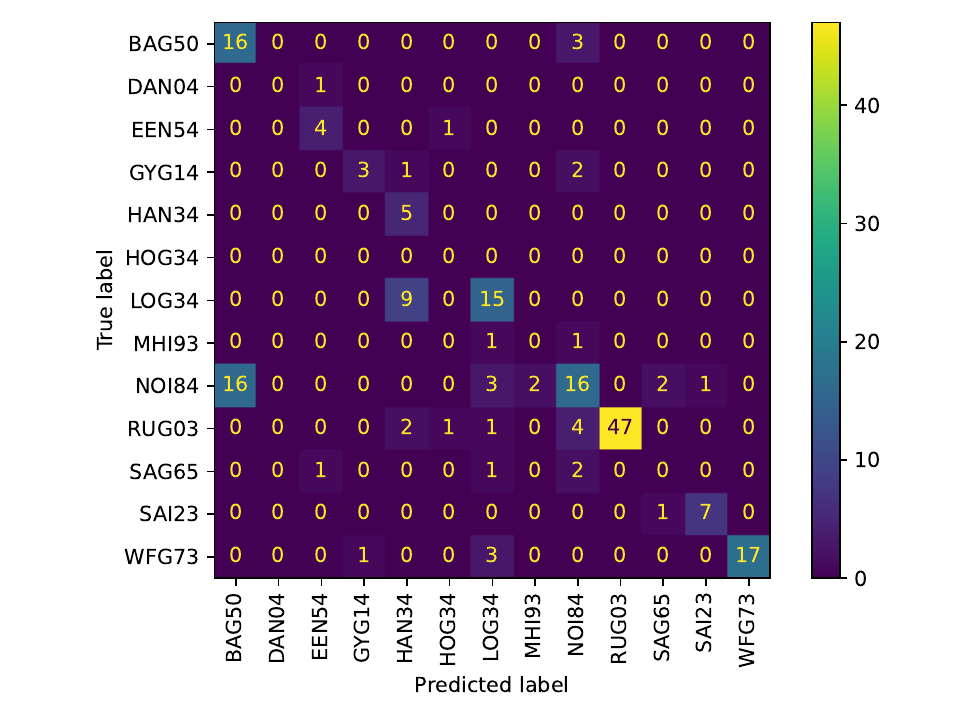}
        \textbf{(c) Confusion matrix of the random forest classifier for the third classroom.}
    \end{minipage}
    \begin{minipage}{0.49\textwidth}
        \centering
        \includegraphics[width=1\linewidth]{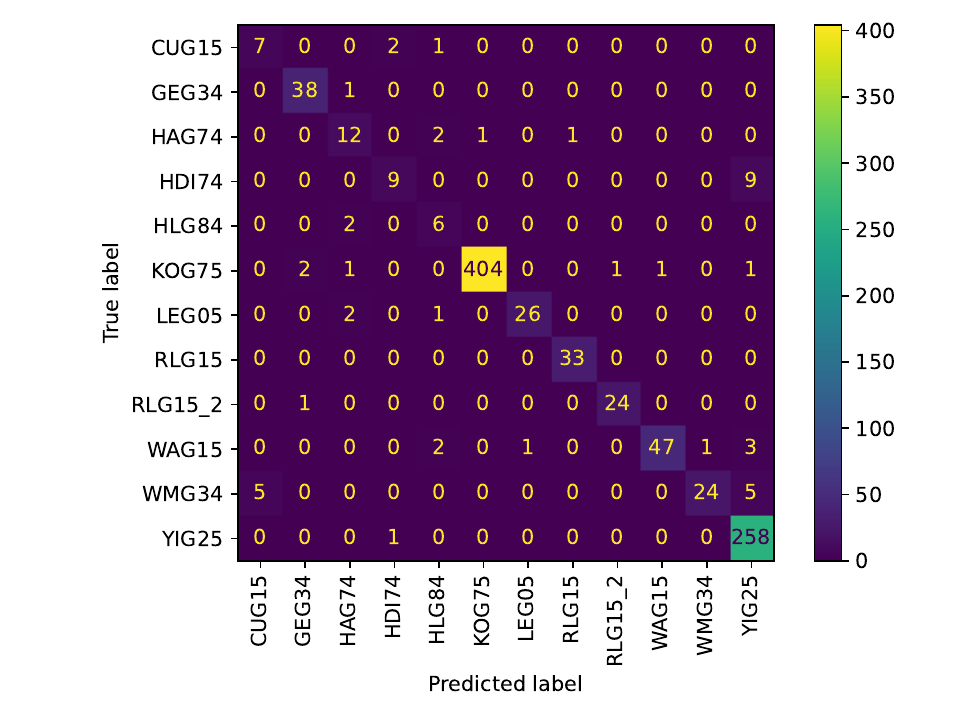}
        \textbf{(d) Confusion matrix of the k-nearest neighbor classifier for the fourth classroom.}
    \end{minipage}
    \caption{Confusion matrices of the best-performing models. Labels correspond to different students in the classrooms.}
    \label{fig_confusion_matrices}
\end{figure*}

We report the results separately as we apply our analyses to each classroom video. For each video, we first report accuracy, precision, recall, and $\text{F}_1$ scores of automated teacher attention mapping results for random forest, support vector machine, k-nearest neighbor, gradient boosting, and decision tree classifiers using ResNet50@WebFace600K and ResNet100@Glint360K facial feature embeddings in Table~\ref{table_stats_results}. For our specific task, random forests, support vector machines, and k-nearest neighbor classifiers have great potential for use in practice. In addition, facial feature embeddings from ResNet50@WebFace600K outperform the baseline from ResNet100@Glint360K. Furthermore, we report confusion matrices for the best-performing models for each classroom video in Figure~\ref{fig_confusion_matrices} to show how the samples for each student correspond to the overall performance. 

As depicted in Table~\ref{table_stats_results}, random forest classifiers outperform the others in almost all classroom setups, namely support vector machines, k-nearest neighbors, gradient boosting, and decision trees. Comparing the k-nearest neighbors and support vector machines with random forests, while there is a directly observable performance difference in the first and second classrooms, the performance of these classifiers is comparable in the third and fourth classrooms, and they are greater than the performance from the first two classrooms. When the number of students and seating arrangements are considered in each of these classrooms, as depicted in Figure~\ref{fig_classroom_screenshots}, fewer students exist in the third and fourth classrooms compared to the first two. In addition, the third classroom's seating structure is U-type, whereas the fourth classroom is smaller in terms of number of students, and the teacher has a better vision of students due to the proximity. However, the first two classrooms are rather conventional, and especially for the faces of the students, a lot of occlusion situations happen. Considering all the results, we found that our automated processing pipeline works accurately when the faces of the students are visible without too many occlusions. 

In addition to the performance analyses, we conducted more fine-grained analyses considering the results for each visually attended student individually by using the precision metric and the confusion matrices provided in Figure~\ref{fig_confusion_matrices}. We found that in the first classroom, the most successful attention mapping results came from one of the students who sat in the front (i.e., OOU86), whereas the second and third best results came from students who sat in the third row (i.e., TAE96 and FLE76). In the second classroom, there is a more apparent trend that the first and second most successful attention mapping results came from the students who were seated in the front part of the classroom (i.e., PIR92 and KAN63), whereas the third most successful recognition results came from a student who sat in the second row (i.e., DEN41).

When we conducted the same analyses for the third and fourth classrooms, we also observed a clear trend, similar to the results for the second classroom. For instance, for the third classroom, the first and the third most successful attention-mapping results came from the students who sat in the frontal part of the classroom (i.e., RUG03 and SAI23), whereas the second most performant results came from the students who were sitting in the visible parts of the classroom (i.e., WFG73). Similarly, the samples of the front-sitting students (e.g., KOG75, WAG15, and RLG15) in the fourth classroom contributed the most to the success of our approach, indicating the importance of the visibility of the student faces for our automated attention mapping pipeline. 

Lastly, across all the classroom scenarios, we found that experiments that utilized facial feature embeddings from the ResNet50@WebFace600K outperformed those that utilized the ResNet100@Glint360K, which we explored as a baseline. WebFace600K~\cite{zhu2021webface260m} dataset applies an automated cleaning pipeline for filtering the high-quality images, whereas Glint360K~\cite{an2021partialfctraining10} is considered one of the cleanest datasets in the face recognition area. As the face patches we extracted from our classroom videos can be classified as noisy images, it is reasonable that facial feature embeddings from the ResNet50@WebFace600K work more accurately than our baseline. 

\section{Discussion}
In eye-tracking research, the post-processing step of mapping AOIs is essential in correlating eye movement metrics with specific parts of the stimulus, such as the duration of gaze on a particular object within that stimulus. The analysis of human attention in the wild, as captured by MET, is a time-consuming process requiring detailed mapping of AOIs in each video frame. The primary objective of this study was to develop a processing pipeline to automate this manual process using computer vision and machine learning techniques. While previous studies have primarily focused on the automated detection of object classes like vehicles~\cite{vasu2021vehicle} or animals~\cite{buehler_2019}, only a few have addressed the complexities of human-to-human interactions, where face recognition becomes pivotal for automated AOI mapping~\cite{Sumer_2018_CVPR_Workshops}. We particularly designed our processing pipeline to efficiently map AOIs in egocentric videos from an eye tracker, focusing on sequences with (multiple) faces. Our goal is to assist researchers who examine human-to-human attention dynamics in natural, real-world settings, with a special focus on challenging environments. To this end, we have pinpointed traditional classroom settings in schools as an ideal context for testing this automated mapping. Classroom environments typically feature dense and complex social interactions, making them well-suited for advancing our understanding of attentional behaviors in situ. The key result of this study indicates that our approach can accurately detect the student receiving the teacher's visual attention across various classroom settings, particularly when employing techniques like random forests and k-nearest neighbors, in conjunction with SCRFD-10GF~\cite{guo2022sample} and ResNet50@WebFace600K~\cite{zhu2021webface260m}. This finding and our processing pipeline can facilitate follow-up work on studying teacher-student interactions, assessing teaching quality, and understanding student learning. 

\subsection{Implications for Automated Attention Mapping in Educational Research}
Our automated teacher attention mapping solution provides many opportunities for educational data analysis and mining, as if it is deployed in the real world; it only requires several face images for each student, just like modern smartphone or tablet facial authentication processes. Once these are in place and the machine learning classifiers are trained, teachers' valid eye-tracking signal on the video stimulus is adequate for its use. Previous research~\cite{buehler_etal_2023} empirically showed that the correlations between automated processes such as hand-raising detection and learning-related student behaviors are comparable to those obtained from the manually labeled data. While we did not explicitly analyze similar correlations with and without using our automated attention mapping pipeline using the teachers' gaze, as our method works accurately across different classrooms, educational researchers can likely utilize our processing pipeline for their research, similar to the previous work, to understand how the teachers' attention is related to different student characteristics and whether manual and automated approaches provide correlated results. 

While we utilized such a processing pipeline for the first time in the literature using the videos from conventional classrooms, previous research focused on smaller and standardized seminar-like classroom settings where the number of students is few, and it is less likely to have significant issues related to occlusions~\cite{Sumer_2018_CVPR_Workshops}. Similar to previous research, we also show that teachers' attention could be accurately mapped even when a greater number of students and issues regarding illuminations and partial occlusions exist. In addition, we found that random forests, k-nearest neighbors, and support vector machines perform well for such purposes, the latter overlapping with the findings of the previous work~\cite{Sumer_2018_CVPR_Workshops}. One of the main caveats is that our automated pipeline works more accurately if students' faces are more visible in the classroom, because it depends on face detection and recognition accuracy (i.e., third and fourth classrooms). Therefore, it is more reasonable to utilize it in classrooms where students' faces are visible during the classroom discourse, especially if researchers plan any further analyses based on the results of our automated pipeline. 

Despite the decent performance, our approach should not be treated as a classroom monitoring system. Due to ethical and privacy reasons, we argue that practitioners should utilize it in real time to avoid recording any personally identifiable image or video data. When it is used in real time, apart from the benefits from manual data annotations point of view, one can also implement additional systems highlighting specific regions in the classrooms with visual or audial cues, complementing our automated pipeline so that if a teacher does not pay attention to the particular student(s) throughout the lesson, such a feedback system can warn teacher about this to help improve teaching quality and further support developing students' self-concepts. 

\subsection{Ethical Considerations}
We process video and eye-tracking data in our methods, and as video data includes the faces of the minors, which are directly identifiable data, ethical considerations are necessary. In addition, it is possible to extract sensitive personal information using eye-tracking data in different setups~\cite{liebling_and_preibusch_2014,bozkir_etal_2023eyetracked}. Therefore, responsible data processing from the teachers' eye-tracking data perspective is also essential. 

As we utilize face detection and recognition models capable of running in real time, when our method is employed, it is possible to process such data without recording it in the first place, ensuring data privacy to a significant extent. In case such video recordings are stored for further offline analyses, anonymization techniques from the classroom context~\cite{suemer_2020_automated} would not only be suitable but also fit well with our approach, as such methods directly work at the face patch level without distorting the remaining parts of the video data. Comparably, as almost all commercial eye-tracking devices support real-time gaze processing, we encourage practitioners to apply these analyses in real time. If storing or sharing eye-tracking data is necessary, especially in the classroom context, computational approaches that can help preserve data privacy should be in place~\cite{bozkir_etal_2020_etra,bozkir_2021_diff_privacy,ozdel_etal_2024}. 

\subsection{Limitations and Future Works}
Our approach has great potential in automating teacher attention mapping processes in classrooms; however, we identified certain limitations when practitioners actively utilize such a pipeline. Firstly, we used the videos from the first two minutes of the classroom instruction to design and assess our method. While such an amount of data is sufficient as it already includes video frames and eye-tracking data from different angles in the classroom due to the physical movement of the teacher, as this data corresponds to the beginning of the recordings, the likelihood of having drifted eye-tracking signal is low as the eye tracker is newly calibrated at the beginning of the data collection. Therefore, if our approach is in place for long durations, it is essential for future works to keep an eye on the additional time fractions regarding the eye-tracking data quality to avoid any inaccuracies in attention mapping. For data quality purposes, teachers can recalibrate the device after a certain time threshold to keep the attention mapping accuracy high. In addition, wearing such an eye tracker might also negatively affect the ecological validity, especially if students and teachers are not familiar with classroom environments with such intelligent systems. However, improving the technology literacy of teachers and students may help with this, especially since it is already needed, considering the fast-growing landscape of machine learning and artificial intelligence.

In our work, we attempted to standardize the classroom situation by using the first minutes of a lesson during which the teacher conducts whole-group instruction, meaning they stand in front of the students, and each student is mainly looking at the teacher. This standardization helps to maintain a consistent and clear view of both the teacher and the students, optimizing the performance of our system. However, in other pedagogical practices like group work, where teachers walk through the room, or students look away more often, the performance of our system might decrease. These dynamic and less structured interactions challenge our current setup as face detection and recognition become more complex with increased movement and significant occlusions. Future works should address these scenarios to ensure the system's robustness across various teaching methods. It may be interesting to dive deeper into the relationship with occlusions and different video segments; however, complete occlusions likely lead to faces not being detected at all. Therefore, either manual occlusion annotations or student detections using body movements may be needed to analyze such a relationship. 

The performance of our automated approach heavily depends on face detection and face recognition results. While we utilized SCRFD-10GF and ResNet50@WebFace600K due to their efficiency and real-time capabilities for our pipeline and ResNet100@Glint360K for baseline comparisons, other models can be interchangeably used. However, if students' faces are occluded due to their seating positions in the classroom or if there are significant illumination changes in the environment, the performance of our pipeline would likely drop. Due to these, we deliberately evaluated our approach when a face or faces are detected in the scene to avoid any performance decrease due to any issue in the face detection task. Future research may especially focus on the relationship between seating positions and classification accuracy for each student, especially when the sample size is bigger than ours. 

Furthermore, researchers following our approach should be aware that the stimulus material to be automatically mapped must include all faces visible for the ground truth annotations. If data from any individual is unavailable (e.g., due to the face detector missing the patches) in different parts of the egocentric video material (i.e., training or test data for the particular machine learning task), one might consider manually including these data samples. Otherwise, these individuals will not be included in the training and testing phases of the experiments, as we also observed in our results. Especially when the training dataset misses some of the individuals, the visual attention mapping performance will automatically drop if these particular individuals are available in the test set. 

Lastly, despite the reasonable performance, we did not utilize the proposed processing pipeline to study student-teacher interactions, teaching quality, or student learning. These remain the focus of our future research. In addition, future research can build on our findings to develop real-time dashboards that inform teachers, visualize engagement levels, highlight areas needing attention, and provide data-driven recommendations for instructional strategies. 

\section{Conclusion}
\label{sec_Conclusion}
In this paper, we proposed an automated attention mapping pipeline to identify which student teachers focus on without needing a vast amount of manually labeled data in eye-tracked classroom setups by using and extending state-of-the-art face detection, face recognition, and gaze tracking methods. Our findings imply that when students' faces are not completely occluded and visible from the teachers' field of view, it is possible to use such an automated pipeline to recognize the visually attended students accurately. As our approach requires minimal manually labeled classroom data, it will likely enrich further analyses in classrooms and educational contexts, such as relationships between teachers' visual attention and different learner profiles and characteristics. In future work, we plan to evaluate our approach in more diverse classrooms with varying seating arrangements and study the relationship between teachers' predicted visual attention and teaching effectiveness.

\section*{Acknowledgments}
Parts of this research were conducted while E.B. and E.K. were affiliated with the University of Tübingen. This research was partly funded by the Deutsche Forschungsgemeinschaft (DFG, German Research Foundation) under Germany’s Excellence Strategy - EXC number 2064/1 - Project number 390727645.
%
\bibliographystyle{abbrv}
\bibliography{sigproc}  
%

\balancecolumns
\end{document}